\documentclass{article} % For LaTeX2e
\usepackage[final]{colm2026_conference}

\usepackage{microtype}
\usepackage{url}
\usepackage{booktabs}
\usepackage{amsmath,amssymb}
\usepackage{graphicx}
\usepackage{xcolor}
\usepackage{listings}
\usepackage{hyperref}
\newcommand{\method}{\textsc{SinkFlex-RL}}
\newcommand{\tautwo}{$\tau^2$-Bench}

\title{Efficient Reinforcement Learning for Long-Horizon Tool-Use Agentic Tasks}

\author{
  \begin{tabular}{c}
    \begin{tabular}{cc}
      Zelei Cheng\thanks{Equal contribution.} & Amritansh Mishra\footnotemark[1] \\[1ex]
      Sambit Sahu & William Campbell 
    \end{tabular} \\[2ex]
    \normalfont AI Foundations, Capital One \\[0.5ex] % \normalfont removes the bolding
    \normalfont \{zelei.cheng, amritansh.mishra, sambit.sahu, william.campbell\}@capitalone.com
  \end{tabular}
}

\begin{document}

\maketitle

\begin{abstract}
Long-horizon tool-using agents must reason over user goals, domain policies, tool calls, simulator state, and delayed verifiable rewards. Reinforcement learning (RL) is a natural fit for this setting, but multi-turn on-policy rollouts create long contexts, while model-specific attention layers may require custom masks and learned sink normalization. We present \method, a modular training system for RL in dual-control tool-use environments. The system combines a Gymnasium-compatible environment wrapper, a VERL-style rollout dataflow, group-relative policy optimization without a separate value model, and a sink-aware FlexAttention path designed to preserve model-specific sink scaling under causal and sliding-window masks. In a preliminary \tautwo{} retail run, validation reward (mean@1) rises from $0.25$ early in training to $0.44$ later in the observed training window, while training-score and trajectory-reward proxies also trend upward. In a fixed-configuration memory benchmark, the optimized attention path reduces peak VRAM from $28.06$~GB to $22.52$~GB at $4096$ tokens, a $19.7\%$ reduction, and runs the measured $8192$-token configuration using $25.53$~GB where the eager baseline runs out of memory. These results illustrate the value of integrating environment interfaces, RL dataflow, and attention-kernel design for memory-feasible long-horizon agent training. 
\end{abstract}

\section{Introduction}

Language-model agents increasingly operate in interactive environments: they converse with users, inspect and update state through tools, follow domain policies, and receive sparse success signals only after long trajectories. Benchmarks such as $\tau$-Bench and $\tau^2$-Bench formalize this setting by evaluating agents in customer-service-style domains with tool APIs, policy constraints, and simulated users~\citep{yao2024taubench,barres2025tau2bench}. These settings stress both reasoning quality and training-system capacity. A single trajectory may contain many dialogue turns, tool calls, and environment observations; on-policy RL multiplies this cost by requiring fresh rollouts, reward checking, and repeated policy updates.

This paper studies the systems side of memory-feasible agentic RL for a large open-weight mixture-of-experts (MoE) transformer. MoE architectures can increase model capacity at lower per-token feed-forward cost by routing tokens to a subset of experts~\citep{fedus2022switch}, but they do not remove the long-context attention bottleneck. As multi-turn trajectories grow, eager attention can exhaust high-bandwidth memory before backpropagation. Our current system evaluation covers sequence lengths through $8192$ tokens; longer target workloads motivate the design but are not evaluated here. Existing fused attention kernels improve memory movement and throughput~\citep{dao2022flashattention,dao2023flashattention2}, but production models may require attention variants with learned sink parameters, heterogeneous masks, or backward behavior that is not exposed by a fixed kernel interface.

We therefore treat agentic RL as a systems-integration problem and make three contributions. First, we wrap dual-control environments behind a Gymnasium-style interface so that rollouts, tools, user simulators, and reward checkers can connect to a standard RL dataflow. Second, we use group-relative policy optimization (GRPO) to update the actor from groups of trajectories without training a separate value model, following the memory-motivated rationale introduced in DeepSeekMath~\citep{shao2024deepseekmath} and a VERL-style post-training dataflow~\citep{sheng2024hybridflow}. Third, we implement a sink-aware FlexAttention path that composes causal and sliding-window masks with differentiable sink scaling while avoiding eager attention-state materialization. We evaluate the integrated pipeline with a preliminary \tautwo{} retail training run and a peak-memory study through $8192$ tokens. The core contribution of this work is the integration of these components and the resulting memory feasibility.

\section{Related Work}

\textbf{Agentic benchmarks.}
Evaluating language models as interactive agents requires environments in which models issue API calls, manage changing state, and process multi-turn feedback. AgentBench~\citep{liu2024agentbench} motivated broad interactive evaluation, while WebArena~\citep{zhou2024webarena}, GAIA~\citep{mialon2024gaia}, and SWE-bench~\citep{jimenez2024swe} target web navigation, general assistant capabilities, and software engineering, respectively. For stateful tool use, $\tau$-Bench evaluates executable state changes rather than text-only judgments~\citep{yao2024taubench}. We use $\tau^2$-Bench~\citep{barres2025tau2bench}, whose dual-control setting allows both the agent and a simulated user to influence task progress. This provides a representative testbed for gathering missing information, recovering from invalid actions, and maintaining consistency over multi-turn interactions.

\textbf{RL fine-tuning for multi-step tool use.}
In complex tool-use environments, annotating every intermediate action is costly, which motivates training with programmatically verifiable episode-level rewards~\citep{zheng2025deepresearcher,dong2026visual}. PPO~\citep{schulman2017ppo} commonly uses a learned value function, adding memory and compute overhead for large language models. GRPO~\citep{shao2024deepseekmath} removes the separate value model and normalizes rewards within groups of sampled outputs. Frameworks such as \texttt{verl}~\citep{sheng2024hybridflow} organize rollout generation, reward computation, and policy updates as a distributed post-training dataflow. Our work adopts these algorithmic and systems components and focuses on integrating them with long-context, model-compatible attention for multi-turn tool-use rollouts.

\textbf{Attention mechanisms for long-context training.}
Transformers use attention as their core sequence-mixing operation~\citep{vaswani2017attention}, but dense attention has quadratic memory and compute in sequence length. FlashAttention and FlashAttention-2 reduce memory traffic by tiling exact attention and avoiding materialized attention matrices~\citep{dao2022flashattention,dao2023flashattention2}. StreamingLLM shows that retaining attention sinks can help stabilize windowed long-context inference~\citep{xiao2023streamingllm}. PyTorch FlexAttention exposes a programmable interface for composing masks and score modifications~\citep{pytorchflexattention}. \method{} combines these ideas by treating efficient attention as both a memory problem and a model-compatibility problem when sink-aware normalization and heterogeneous masks must remain in the differentiable training path.

\section{Agentic RL Setting}

\paragraph{Dual-control environment.}
In single-control tool-use benchmarks, the agent is often the only actor that changes the environment through tools while the user supplies information passively. In dual-control environments, both sides can affect the world state: the agent chooses dialogue and API actions, while the simulated user can confirm choices, provide missing information, or take user-side actions. A task instance can be represented as
\begin{equation}
  \mathcal{E} = (g, \pi_{\mathrm{SOP}}, \mathcal{A}_{\mathrm{tool}}, s_0, u),
\end{equation}
where $g$ is the user goal, $\pi_{\mathrm{SOP}}$ denotes domain-policy or standard-operating-procedure constraints, $\mathcal{A}_{\mathrm{tool}}$ is the agent tool set, $s_0$ is the initial shared state, and $u$ is the user simulator. A trajectory is
\begin{equation}
  \tau = \{(o_t, a_t, r_t, d_t, i_t)\}_{t=1}^{T},
\end{equation}
where $o_t$ is an observation, $a_t$ is an agent action, $r_t$ is a reward or diagnostic signal, $d_t$ is a termination flag, and $i_t$ contains metadata. The trainer consumes a trajectory-level reward $R_i$ produced by the benchmark's programmatic checker. Other dashboard diagnostics are logged separately and are not outputs from a separately trained critic network.

\paragraph{Why this setting stresses the training system.}
The environment creates three systems pressures. First, rewards are delayed and often verifiable only after the final state is checked, increasing the number of sampled tokens per useful gradient. Second, trajectories are multi-turn and tool-heavy, so the policy must retain user messages, tool outputs, and domain-policy constraints in context. Third, the system must coordinate environment execution, rollout generation, reward checking, and policy optimization while maintaining consistent episode state and policy versions.

\begin{figure}[t]
\centering
\includegraphics[width=\textwidth]{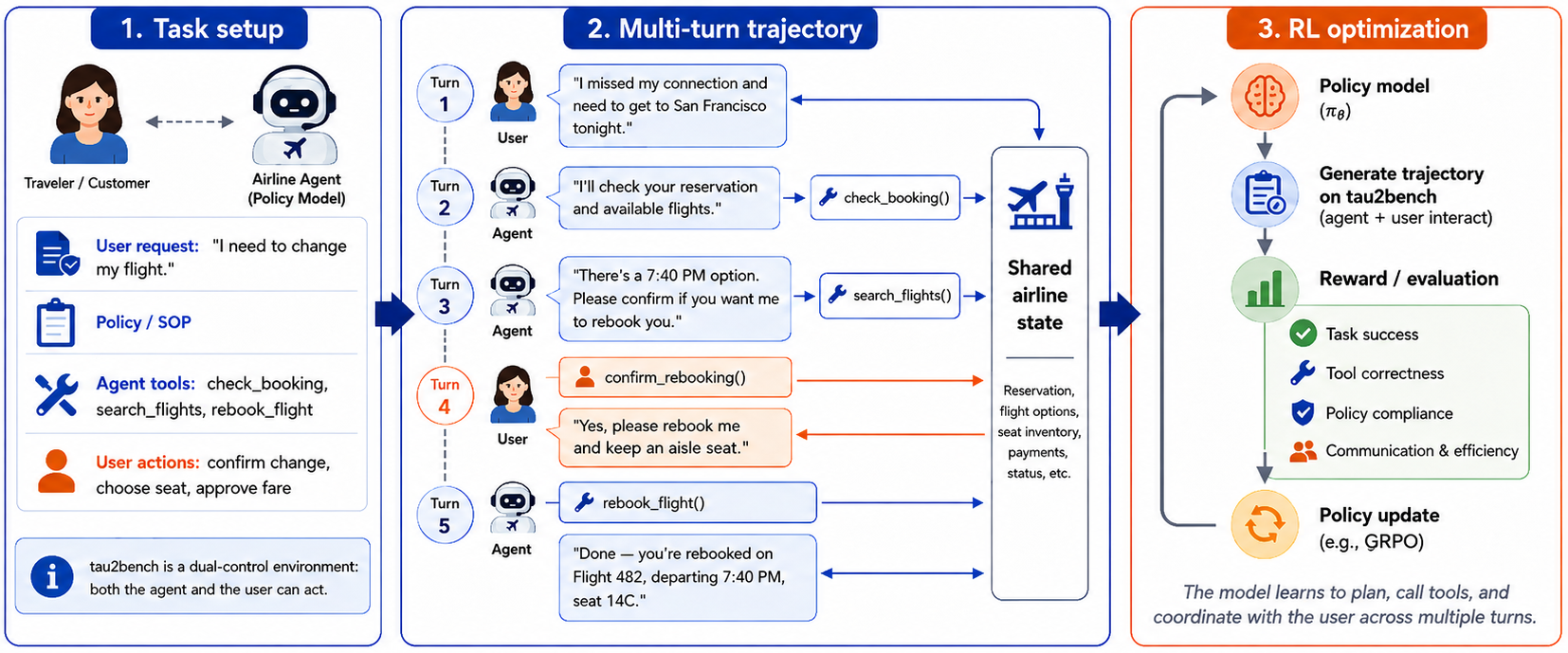}
\caption{A dual-control agentic RL episode. A task specification initializes the environment, user--agent interaction produces a multi-turn trajectory, and a programmatic checker scores the resulting rollout group for a GRPO-style policy update.}
\label{fig:agentic_task}
\end{figure}

Figure~\ref{fig:agentic_task} summarizes the dual-control task structure used in this work: a task specification initializes a shared world state, the user and agent take alternating actions, and the resulting trajectory is scored by a programmatic checker before the rollout group is used for a GRPO-style update.

\section{Training Pipeline}

\begin{figure}[t]
\centering
\resizebox{\linewidth}{!}{%
\includegraphics{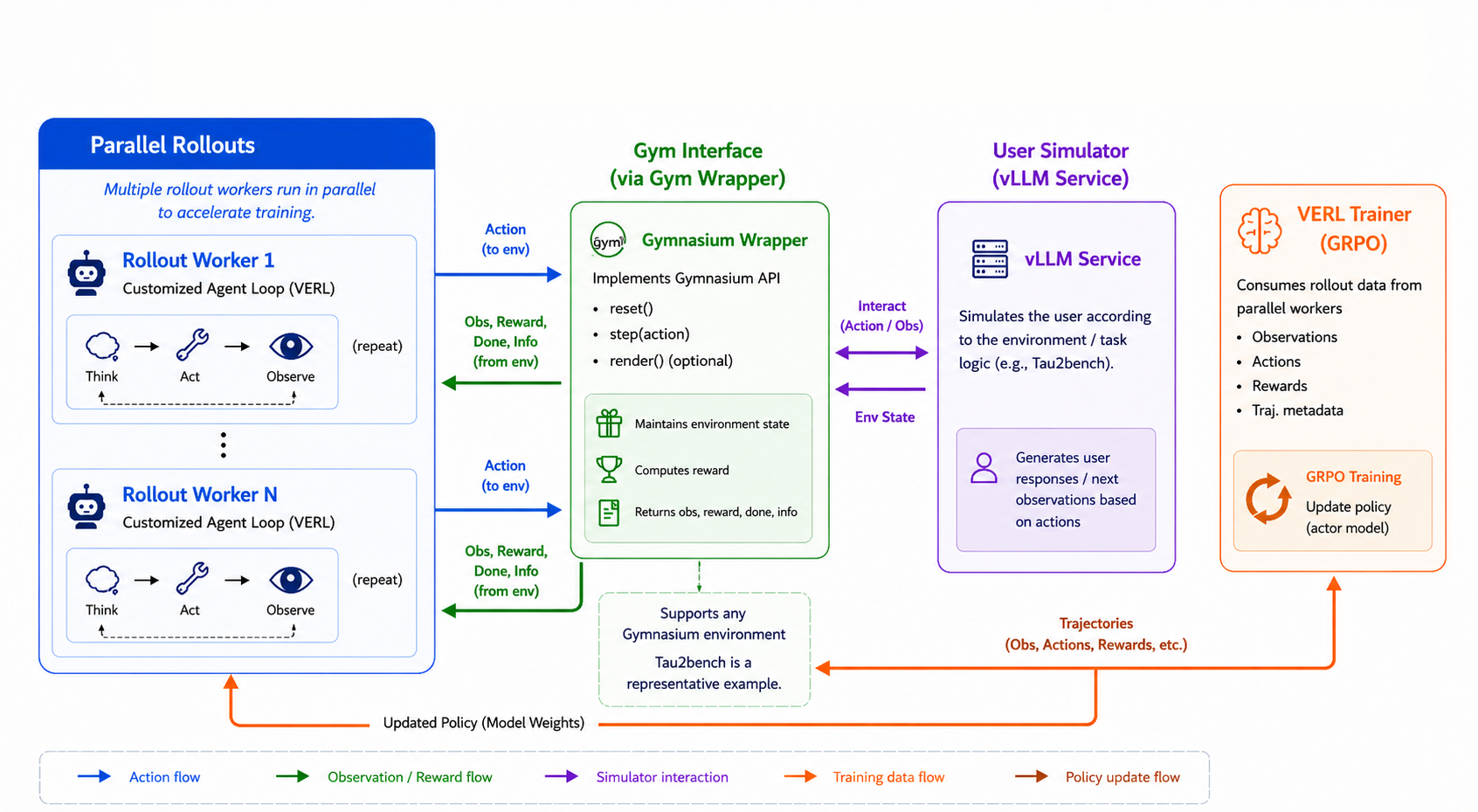}
}
\caption{Training-system dataflow. The actor samples multi-turn trajectories through rollout workers and a Gymnasium-compatible wrapper; trajectory rewards and metadata flow to a GRPO trainer, which updates and redistributes policy weights.}
\label{fig:pipeline}
\end{figure}

Figure~\ref{fig:pipeline} summarizes the training dataflow. We describe the components abstractly to avoid dependence on deployment-specific infrastructure.

\paragraph{Environment wrapper.}
Each benchmark domain is exposed through a common reset/step interface. On reset, the wrapper samples a task, initializes the shared state, and constructs the first policy observation. On step, it routes model outputs to one of three handlers: natural-language response, tool call, or termination. The wrapper then advances the user simulator and state backend and returns the next observation together with reward metadata. This separation keeps benchmark-specific user simulation, tool behavior, and checking logic outside the trainer and is intended to support additional executable environments through the same interface.

\paragraph{Rollout worker.}
The rollout worker owns the agent loop. At each turn, it formats the observation, samples from the current policy $\pi_{\theta}$, parses the response into an action, and appends the transition to the trajectory buffer. The worker repeats until the environment terminates or a maximum-turn budget is reached. The trainer receives token-level log probabilities, action masks, trajectory rewards, and episode metadata without directly depending on benchmark-specific parsing or tool implementations.

\paragraph{Policy update.}
Given a prompt or task context $x$ and a group of $G$ sampled rollouts $y_1,\ldots,y_G$ from the old policy, GRPO computes a group-normalized trajectory advantage
\begin{equation}
  \hat{A}_i = \frac{R_i - \mu(R_{1:G})}{\sigma(R_{1:G}) + \epsilon_A},
\end{equation}
where $R_i$ is the programmatically computed trajectory reward and $\epsilon_A$ is a small numerical constant. Under outcome-level supervision, the same normalized advantage is assigned to every optimized token in rollout $y_i$. For token $y_{i,t}$, the importance-sampling ratio is
\begin{equation}
  \rho_{i,t}(\theta)
  = \frac{\pi_{\theta}(y_{i,t}\mid x,y_{i,<t})}
         {\pi_{\theta_{\mathrm{old}}}(y_{i,t}\mid x,y_{i,<t})}.
\end{equation}
Define the clipped ratio
\begin{equation}
  \bar{\rho}_{i,t}(\theta)
  = \operatorname{clip}\!\left(
      \rho_{i,t}(\theta),1-\epsilon_c,1+\epsilon_c
    \right).
\end{equation}
The loss minimized by the trainer is
\begin{align}
  \mathcal{J}_{i,t}(\theta)
  &= \min\!\left(
      \rho_{i,t}(\theta)\hat{A}_i,
      \bar{\rho}_{i,t}(\theta)\hat{A}_i
    \right), \\
  \mathcal{L}_{\mathrm{GRPO}}(\theta)
  &= -\frac{1}{\sum_i T_i}
     \sum_{i=1}^{G}\sum_{t=1}^{T_i}\mathcal{J}_{i,t}(\theta)
     + \beta D_{\mathrm{KL}}(\pi_\theta\|\pi_{\mathrm{ref}}),
\end{align}
where $T_i$ is the optimized token length of rollout $i$, $\epsilon_c$ is the clipping radius, and the KL term regularizes the policy toward the reference model. This is the minimization form of the usual PPO/GRPO clipped maximization objective~\citep{schulman2017ppo}. The baseline design does not train a separate critic or value network.

\section{Sink-Aware FlexAttention}

\paragraph{Memory wall.}
A trajectory of length $n$ induces an $n\times n$ attention score structure for each relevant batch/head component when implemented eagerly. The number of score positions therefore grows quadratically with context length; for example, $n=20{,}000$ corresponds to $4\times10^8$ positions before accounting for activations, optimizer state, MoE routing, or rollout batching. Fused attention methods address this by computing attention in tiles and reducing high-bandwidth-memory traffic~\citep{dao2022flashattention}. A model-specific attention layer, however, may also require custom score modifications, mixed causal and sliding-window masks, or learned sink logic.

\paragraph{Attention sinks.}
Attention sinks are tokens or learned mechanisms that absorb attention mass and can stabilize long-context behavior. Prior work shows that retaining initial sink tokens can recover quality under windowed attention in streaming inference~\citep{xiao2023streamingllm}. Our setting differs from inference-only KV-cache management: the sink behavior is part of the trainable attention computation and therefore must remain connected to the forward and backward paths.

\paragraph{Zero-value-sink equivalence.}
To avoid materializing an explicit sink token in the key and value caches, we use the algebraic form of a zero-valued sink. Let $s_{\eta}$ be a learned sink logit and let its value vector be $v_{\mathrm{sink}}=\mathbf{0}$. For a query $q$, the attention output with the explicit sink is
\begin{align}
O_{\mathrm{sink}}
&= \frac{\sum_i \exp(q\cdot k_i)v_i + \exp(s_{\eta})\mathbf{0}}
        {\sum_i \exp(q\cdot k_i)+\exp(s_{\eta})} \\
&= \left(
\frac{\sum_i \exp(q\cdot k_i)}
     {\sum_i \exp(q\cdot k_i)+\exp(s_{\eta})}
\right) O_{\mathrm{std}}.
\end{align}
Writing $\ell=\log\sum_i\exp(q\cdot k_i)$ gives
\begin{equation}
  \alpha_{\mathrm{sink}}
  = \frac{\exp(\ell)}{\exp(\ell)+\exp(s_{\eta})}
  = \sigma(\ell-s_{\eta}),
  \qquad
  O_{\mathrm{sink}}=\alpha_{\mathrm{sink}}O_{\mathrm{std}}.
\end{equation}
Under the zero-value assumption, explicit sink materialization is therefore algebraically equivalent to scaling the standard attention output by a factor computed from the log-sum-exp statistic.

\begin{figure}[ht]
\centering
\begin{lstlisting}[language=Python, basicstyle=\ttfamily\scriptsize,
  keywordstyle=\color{blue}, commentstyle=\color{gray}, frame=single]
import torch
import torch.nn.functional as F

# 1. Explicit zero-value sink materialization
scores_with_sink = torch.cat([sink_score, scores], dim=-1)
v_with_sink = torch.cat([torch.zeros_like(v[:, :, :1, :]), v], dim=-2)
out_explicit = F.softmax(scores_with_sink, dim=-1) @ v_with_sink

# 2. Equivalent output scaling
out_std = F.softmax(scores, dim=-1) @ v
lse = torch.logsumexp(scores, dim=-1, keepdim=True)
out_scaled = out_std * torch.sigmoid(lse - sink_score)

assert torch.allclose(out_explicit, out_scaled, atol=1e-6)
\end{lstlisting}
\caption{Illustration of the zero-value-sink identity. The production implementation uses fused attention statistics rather than materializing the dense score tensor shown in this small reference example.}
\label{fig:sink_proof}
\end{figure}

\paragraph{Implementation pattern.}
The implementation uses FlexAttention as a programmable attention substrate~\citep{pytorchflexattention}. A mask function composes causal constraints with an optional sliding-window constraint and an always-visible prefix. The mask is compiled into a block-sparse structure so that fully masked blocks can be skipped. When requested, the attention call returns both the output and an auxiliary log-sum-exp statistic. The sink path then applies the model-specific scaling function to the output:
\begin{align}
  M_{b,h,q,k}
  &= \mathbb{1}[k\le q]
     \wedge
     \mathbb{1}[q-k\le w\;\vee\;k<p], \\
  (z,\ell)
  &= \mathrm{FlexAttention}(Q,K,V;\mathrm{BlockMask}(M)), \\
  \alpha_{\mathrm{sink}}
  &= f_{\eta}(\ell), \\
  z'
  &= z\odot\alpha_{\mathrm{sink}},
\end{align}
where $w$ is the local-window size, $p$ is the number of always-visible prefix positions, and $f_{\eta}$ is the learned sink-scaling function. For the zero-value-sink construction above, $f_{\eta}(\ell)=\sigma(\ell-s_{\eta})$. The prefix budget $p$ and the learned sink logit $s_{\eta}$ are distinct: $p$ controls token visibility in the mask, whereas $s_{\eta}$ reallocates softmax mass through output scaling.

\paragraph{Gradient flow and autograd integration.}
The data-dependent sink scale couples the gradients of the attention output $z$ and the log-sum-exp statistic $\ell$. Let $d$ denote the head dimension. By the chain rule,
\begin{align}
  \nabla_z\mathcal{L}
  &= \nabla_{z'}\mathcal{L}\odot\alpha_{\mathrm{sink}}, \\
  \nabla_{\alpha_{\mathrm{sink}}}\mathcal{L}
  &= \sum_{j=1}^{d}
     \left(\nabla_{z'}\mathcal{L}\odot z\right)_j, \\
  \nabla_{\ell}\mathcal{L}
  &= \nabla_{\alpha_{\mathrm{sink}}}\mathcal{L}
     \odot f_{\eta}'(\ell),
\end{align}
and the sink-parameter gradient is
\begin{equation}
  \nabla_{\eta}\mathcal{L}
  = \nabla_{\alpha_{\mathrm{sink}}}\mathcal{L}
    \odot \frac{\partial f_{\eta}(\ell)}{\partial\eta}.
\end{equation}
A fixed fused-kernel interface may not expose the auxiliary gradient path through $\ell$. Our implementation composes FlexAttention with the sink-scaling operation under AOTAutograd and \texttt{torch.compile}, allowing the compiler to generate forward and backward code for $Q$, $K$, $V$, and the sink parameters without materializing an $O(n^2)$ Jacobian.

\paragraph{Why a programmable kernel interface is needed.}
Optimized attention kernels are essential, but a fixed interface can omit model-specific masking, score modification, or auxiliary-gradient behavior. In on-policy RL, a small attention mismatch can be repeatedly amplified across long trajectories and policy updates. The FlexAttention path therefore prioritizes explicit control over the model's attention semantics while retaining a fused, block-sparse execution path.

\paragraph{Memory-oriented implementation optimizations.}
Profiling identified two avoidable sources of memory allocation: eager materialization in the sink-rescaling computation and replication of mask metadata across batch and head dimensions. We address them as follows.

\paragraph{Compilation and fusion.}
We apply \texttt{torch.compile} to the composed FlexAttention and sink-scaling path. PyTorch Inductor can fuse eligible pointwise operations with surrounding generated code, reducing the number and lifetime of materialized intermediate tensors.

\paragraph{Mask broadcasting.}
When the attention pattern is shared across batch elements and heads, block-mask construction omits explicit batch and head dimensions and relies on kernel-side broadcasting. This avoids storing repeated copies of the same sparse mask metadata. The aggregate effect of the optimized path is evaluated through peak-VRAM measurements in Section~\ref{sec:experiments}; we do not report a separate per-optimization ablation.

\section{Experiments}
\label{sec:experiments}

We report two complementary measurements: preliminary policy-learning trends in the \tautwo{} retail domain and peak-memory scaling for the attention implementation. The first documents behavior over an observed training window; it is not an algorithmic comparison. The second measures peak VRAM only and does not establish throughput or wall-clock improvements.

\subsection{Experimental Design}
\label{subsec:experimental_design}

\paragraph{Experiment 1: Preliminary retail training.}
We evaluate the integrated GRPO pipeline on the retail domain of \tautwo~\citep{barres2025tau2bench}. The agent interacts with a dynamic user simulator, executes multi-step API calls, and follows domain-specific standard operating procedures. The purpose of this experiment is to document whether validation reward and associated training diagnostics improve over the observed run, not to isolate the causal effect of GRPO relative to another optimizer.

We track three metrics:
\begin{itemize}
    \item \textbf{Validation Reward (mean@1):} the benchmark's programmatically verified task-success score for a single sampled validation trajectory per task.
    \item \textbf{Training Score Proxy:} a dashboard diagnostic monitored during training. The dashboard historically labels this trace as a ``critic-style score,'' but the baseline does not train a separate critic or value network.
    \item \textbf{Trajectory Reward Proxy:} a rolling summary of raw episode rewards used to monitor the direction and variability of the observed training run.
\end{itemize}

\paragraph{Experiment 2: Peak-memory scaling.}
We compare the model-native eager attention reference path with the optimized sink-aware FlexAttention path under the same fixed model, batch, and training configuration. The optimized path is configured to reproduce the same causal/sliding-window policy and sink scaling as the reference path. We measure peak high-bandwidth-memory usage (\textbf{Peak VRAM in GB}) at sequence lengths of $1024$, $2048$, $4096$, and $8192$ tokens. Because the current evaluation reports neither throughput nor a forward/backward numerical-equivalence test, the conclusions are limited to observed peak memory and execution feasibility in this configuration.

\subsection{Experimental Results}
\label{subsec:experimental_results}

\subsubsection{Preliminary Retail Training Results}
\label{subsec:performance_results}

Table~\ref{tab:results} summarizes values estimated from the available dashboard screenshots. The retail validation reward rises from $0.25$ early in training to $0.44$ later in the observed training window. These values represent two portions of the same run rather than an untrained baseline and a final converged model.

\begin{table}[htbp!]
\caption{\textbf{Preliminary retail training trends.} Values are visually estimated from dashboard screenshots and should be interpreted as approximate. Higher values are better.}
\label{tab:results}
\begin{center}
\small
\setlength{\tabcolsep}{5pt}
\begin{tabular}{lcc}
\toprule
Metric & Early training & Later training \\
\midrule
Retail validation reward (mean@1) & $0.25$ & $0.44$ \\
Training score proxy (mean) & $0.18$ & $0.40$ \\
Trajectory reward proxy (mean) & $0.18$ & $0.39$ \\
\bottomrule
\end{tabular}
\end{center}
\end{table}

\begin{figure}[ht]
\centering
\begin{minipage}{0.485\linewidth}
\centering
\includegraphics[width=\linewidth]{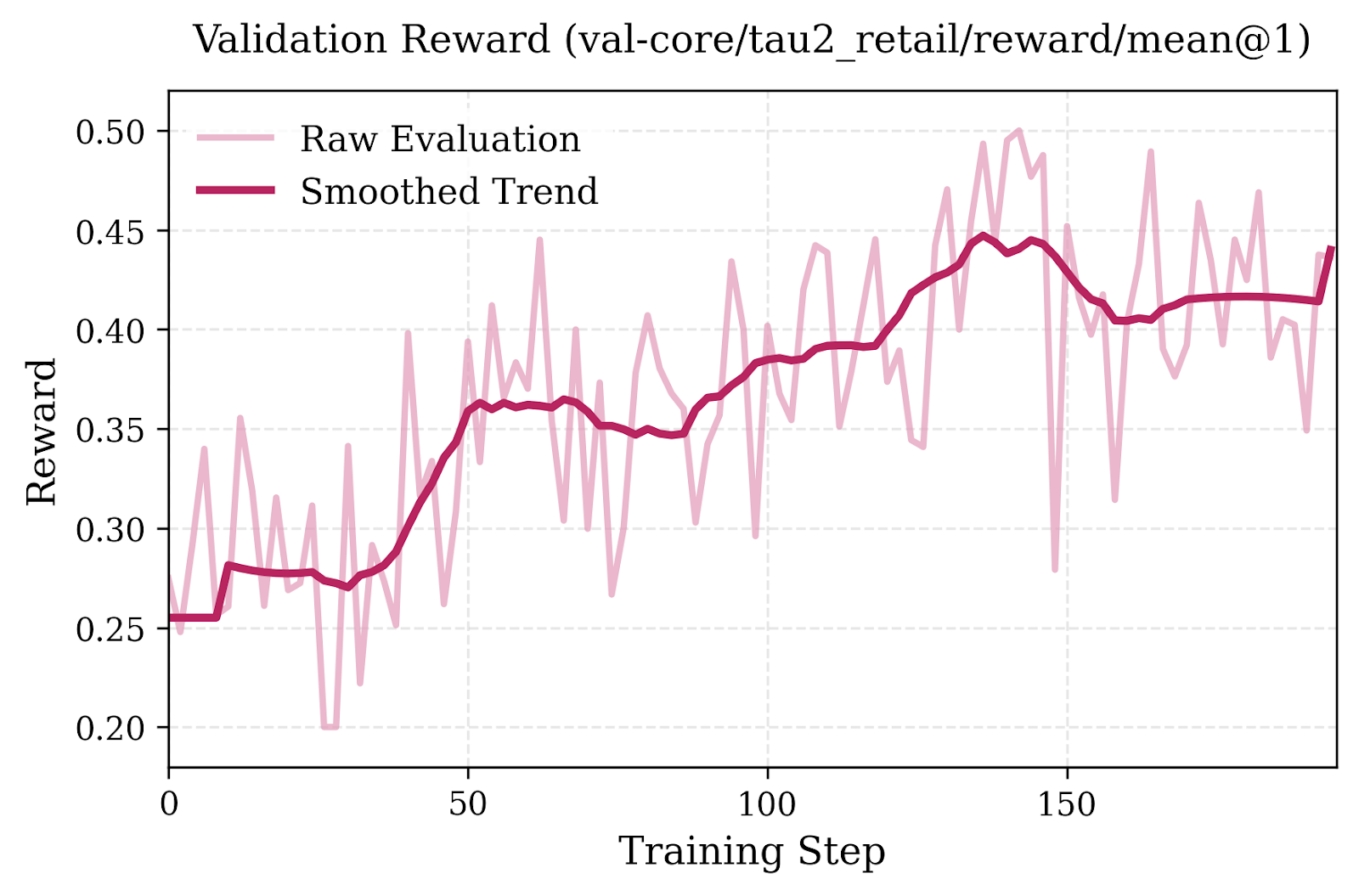}
\end{minipage}\hfill
\begin{minipage}{0.485\linewidth}
\centering
\includegraphics[width=\linewidth]{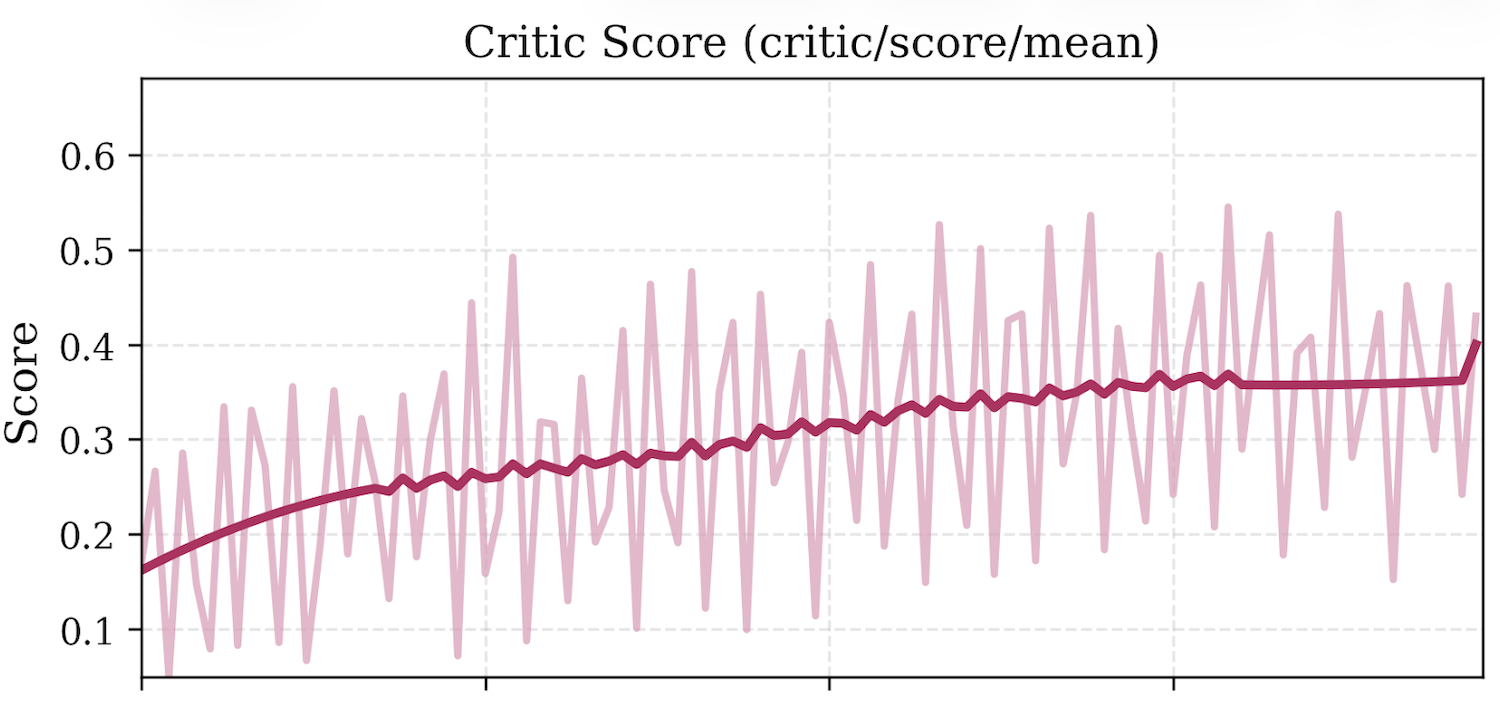}
\end{minipage}
\caption{\textbf{Approximate learning curves.} The left panel shows retail validation reward; the right panel shows the training-score diagnostic whose dashboard label is ``critic-style score.'' Dots denote noisy per-evaluation readings estimated from screenshots, and thick lines show the displayed smoothed trends.}
\label{fig:curves}
\end{figure}

Individual evaluations are variable, as expected for multi-turn, state-dependent tasks. Over the displayed window, the smoothed validation trace trends upward, while the training-score and trajectory-reward proxies rise from $0.18$ to $0.40$ and $0.39$, respectively. These observations are consistent with the integrated pipeline receiving a useful learning signal under the implemented GRPO update. Without multiple seeds, an optimizer baseline, or exported scalar logs, they do not isolate the effect of GRPO, establish variance reduction, or support a statistical significance claim.

\subsubsection{Peak-Memory Results}
\label{subsec:efficiency_results}

Table~\ref{tab:memory_results} reports peak VRAM for the two attention paths. The optimized path uses less memory at every sequence length for which both paths complete, while the eager path runs out of memory at $8192$ tokens.

\begin{table}[htbp]
\caption{\textbf{Peak-memory comparison.} Peak VRAM during the measured training configuration. The optimized sink-aware FlexAttention path completes the $8192$-token configuration for which the eager reference path runs out of memory.}
\label{tab:memory_results}
\begin{center}
\small
\setlength{\tabcolsep}{5pt}
\begin{tabular}{lccc}
\toprule
Sequence Length & Eager Baseline (GB) & Flex + Sink (GB) & Savings \\
\midrule
1024 & 21.07 & 20.11 & 0.96\,GB (4.5\%) \\
2048 & 23.27 & 21.02 & 2.25\,GB (9.7\%) \\
4096 & 28.06 & 22.52 & 5.54\,GB (19.7\%) \\
8192 & OOM   & 25.53 & -- \\
\bottomrule
\end{tabular}
\end{center}
\end{table}

At $4096$ tokens, the optimized path reduces peak VRAM from $28.06$~GB to $22.52$~GB, a reduction of $5.54$~GB or $19.7\%$. At $8192$ tokens, it completes the measured configuration with a peak allocation of $25.53$~GB, whereas the eager reference path encounters an out-of-memory error. Thus, in this fixed configuration, the optimized path removes the $8192$-token memory failure observed for the eager baseline. Because the experiment measures only peak VRAM, it does not establish improvements in throughput, latency, total training time, or accelerator utilization.

\section{Discussion and Limitations}

First, our system is particularly well suited to agentic tasks that operate in controlled environments with programmatically verifiable outcomes. This setting is naturally instantiated by \tautwo, where the policy interacts with a user simulator, invokes domain tools, and receives feedback based on the resulting environment state. By grounding optimization in auditable actions and objective task outcomes, the framework enables scalable and reproducible policy learning. The same approach has strong potential to extend to other domains that provide structured tool interfaces, executable environments, or clearly defined success criteria. For more open-ended applications, programmatic verification can be complemented with human feedback or learned reward models, providing a natural path toward broader task coverage.

Second, the preliminary retail-domain experiment provides encouraging evidence for the effectiveness of the proposed policy-learning pipeline. Over the observed training window, the displayed reward-associated traces exhibit a clear upward trend from the early to later stages of training, indicating that the policy is able to benefit from the optimization signal. These results serve as a promising proof of concept for applying GRPO to tool-using agentic systems. They also motivate broader evaluations across domains, random seeds, model scales, and training horizons to further characterize the robustness and generality of the observed improvements.

Third, programmatic rewards provide an efficient and scalable source of supervision by directly connecting policy behavior to verifiable task outcomes. This foundation can support increasingly rich reward functions that jointly assess final-state correctness, policy compliance, tool-use quality, and dialogue naturalness. GRPO is especially well matched to settings in which rollout groups contain diverse behavioral outcomes, allowing relative advantages to identify and reinforce stronger trajectories. Techniques such as diversity-aware sampling, curriculum design, adaptive grouping, and reward shaping offer promising directions for maintaining informative reward variation and improving optimization efficiency.

Forth, the memory evaluation demonstrates the feasibility of executing the proposed training approach at sequence lengths of up to $8192$ tokens under the tested configuration. In particular, the sink-aware execution path provides a practical foundation for scaling agentic policy learning to longer interaction trajectories while retaining the intended eager reference semantics. The current peak-VRAM measurements establish an important systems proof of concept, and the same evaluation framework can be extended to characterize throughput, latency, training time, and accelerator utilization. Forward- and backward-equivalence tests, together with targeted validation of sink handling, masking, position indexing, and log-sum-exp statistics, can provide additional confirmation of semantic consistency. Overall, these results establish a strong foundation for developing efficient, reliable, and scalable training systems for long-horizon tool-using agents.

\section{Conclusion}

We presented \method, a modular RL training system that integrates a Gymnasium-compatible dual-control environment interface, VERL-style rollout dataflow, GRPO updates without a separate value model, and a sink-aware FlexAttention path. In the current evidence, retail validation and training diagnostics trend upward over one preliminary run, while the optimized attention path reduces peak memory and completes the measured $8192$-token configuration where the eager reference path runs out of memory. These results support the narrower claim that environment, RL-dataflow, and attention-kernel integration can improve the memory feasibility of long-horizon agent training. They do not yet establish algorithmic superiority, broad generalization, exact implementation equivalence, or end-to-end computational speedup.

\section*{Ethics Statement}

The system trains agents that can call tools and update environment state. Such agents should be evaluated for policy compliance, user deception, unsafe tool use, privacy leakage, and simulator overfitting before deployment. The present work focuses on system feasibility and preliminary training behavior and does not claim deployment readiness.

\bibliography{colm2026_conference}
\bibliographystyle{colm2026_conference}

\end{document}